\documentclass[10pt,conference]{IEEEtran}
\usepackage{graphicx}
\usepackage[utf8]{inputenc}
\usepackage{amssymb,amsmath,array}
\usepackage{comment}
\usepackage{booktabs}
\usepackage{epstopdf}
\usepackage[hidelinks]{hyperref}

\usepackage{ifpdf}

\ifpdf
  \DeclareGraphicsExtensions{.pdf,.png,.jpg}
\else
  \DeclareGraphicsExtensions{.eps}
\fi

\usepackage[font=small,labelfont=bf]{caption}
\title{SEAL: Searching Expandable Architectures for Incremental Learning}

\author{\IEEEauthorblockN{Matteo Gambella and Manuel Roveri}
\IEEEauthorblockA{Dipartimento di Elettronica, Informazione e Bioingegneria,\\
Politecnico di Milano, Milan, Italy\\
Email: \{matteo.gambella, manuel.roveri\}@polimi.it} 
}

\begin{document}

\maketitle






\begin{abstract}
Incremental learning is a machine learning paradigm where a model learns from a sequential stream of tasks. This setting poses a key challenge: balancing plasticity (learning new tasks) and stability (preserving past knowledge). Neural Architecture Search (NAS), a branch of AutoML, automates the design of the architecture of Deep Neural Networks and has shown success in static settings. However, existing NAS-based approaches to incremental learning often rely on expanding the model at every task, making them impractical in resource-constrained environments. In this work, we introduce SEAL, a NAS-based framework tailored for data-incremental learning, a scenario where disjoint data samples arrive sequentially and are not stored for future access. SEAL adapts the model structure dynamically by expanding it only when necessary, based on a capacity estimation metric. Stability is preserved through cross-distillation training after each expansion step. The NAS component jointly searches for both the architecture and the optimal expansion policy. Experiments across multiple benchmarks demonstrate that SEAL effectively reduces forgetting and enhances accuracy while allocating additional capacity only when required. These results highlight the promise of combining NAS and selective expansion for efficient, adaptive learning in incremental scenarios.
\end{abstract}



\section{Introduction}
\label{sec:introduction}

Incremental Learning (IL), or Continual Learning, enables models to learn from a sequential stream of tasks~\cite{continual-survey}. In this setting, each task introduces new observations from which the model must learn to perform actions. However, training Deep Neural Networks (DNNs) in this context typically leads to two major challenges: catastrophic forgetting, the tendency to overwrite knowledge from previous tasks~\cite{types-incremental}, and plasticity loss, the model’s inability to learn new information effectively~\cite{plasticiy-loss}. IL methods aim to mitigate these issues by promoting stability (e.g., via regularization~\cite{ewc, si-baseline} or knowledge distillation~\cite{lwf}) and maintaining plasticity (e.g., through dynamic network expansion~\cite{den-continual, adaxpert}). These techniques are especially valuable in resource-constrained scenarios such as IoT and TinyML~\cite{tybox, tinymlwizardry, leafsense, hcvriot}.

Another promising line of work focuses on the geometry of the loss landscape. By guiding optimization toward flatter regions, models become more robust to perturbations~\cite{sam, entropic-flat, fantastic-generalisation}, which is beneficial in IL since new tasks can be viewed as perturbations to previously learned solutions~\cite{plasticity-losslandscape}.

In parallel, Automated Machine Learning (AutoML), particularly Neural Architecture Search (NAS), aims to automate the design of neural networks~\cite{nas_survey_components}. Flatness has emerged as a desirable property in NAS to enhance generalization and robustness~\cite{flatnas, neighborhood-aware, genas, a2m}. NAS has also been applied to IL to adapt architectures across tasks~\cite{nas-continual-learning, cleas, learning-to-grow}, often by searching for optimal expansions. However, most existing approaches repeat the search process at every task, incurring high computational costs. Recent perspectives have argued that future NAS systems should jointly address efficiency, robustness, and continual learning, treating architectural adaptivity as a first-class design objective rather than a post-hoc addition \cite{hercules}.

In this work, we introduce SEAL (Searching Expandable Architectures for Incremental Learning), a novel NAS framework for IL. Unlike prior methods, SEAL jointly searches both the optimal architecture and its expansion policy within a single search phase. This unified approach significantly reduces computational overhead and memory usage while maintaining adaptability across tasks. As a result, SEAL offers a more practical solution for real-world deployment and aligns with key desiderata of NAS for IL, such as efficiency, scalability, and adaptability~\cite{nas-continual-learning}.
SEAL specifically targets data-incremental learning, where tasks correspond to disjoint subsets of data while maintaining the same label space~\cite{types-incremental}. 

SEAL utilizes a similar design to FlatNAS~\cite{flatnas}, using an OFA supernet for faster evaluation
of the models~\cite{ofa} and adaptive switching for surrogate prediction~\cite{nsganetv2}. To accelerate and improve performance estimation, we constrain the expansion operator so that all expanded architectures remain valid OFA subnetworks. This allows fast evaluation via fine-tuning, using pre-trained OFA weights to initialize newly added components.

We evaluate SEAL on three datasets, showing that it delivers competitive performance while selectively allocating additional capacity only when required. Moreover, incorporating a flatness-based robustness metric yields consistent improvements in IL metrics.

The remainder of this paper is structured as follows. Section~\ref{sec:relatedliterature} shows the related literature. Section~\ref{sec:background} introduces relevant background concepts. Section~\ref{sec:seal} details the proposed SEAL framework. Section~\ref{sec:experimental_results} presents experimental results, and Section~\ref{sec:conclusions} concludes the paper. To support reproducibility, we will release SEAL's source code upon paper acceptance.

\section{Related Literature}
\label{sec:relatedliterature}
Several techniques have been proposed to address the challenges of incremental learning and architecture robustness. Among the earliest methods tackling model expansion, \textit{Dynamic Expandable Networks} (DEN)~\cite{den-continual} balance plasticity and stability by enforcing sparse weight usage and selectively expanding the network when a loss threshold is exceeded. \textit{AdaXpert}~\cite{adaxpert} refines this idea by using the Wasserstein distance to measure task similarity and determine when and how much to expand, based on the observed performance gap. Regularization-based methods such as \textit{Elastic Weight Consolidation} (EWC)~\cite{ewc} and \textit{Synaptic Intelligence} (SI)~\cite{si-baseline} mitigate forgetting by penalizing changes to parameters deemed important for previous tasks. \textit{Learning without Forgetting} (LwF)~\cite{lwf} takes a different approach by using knowledge distillation to retain prior knowledge without revisiting old data. More recently, \textit{Tybox}~\cite{tybox} focuses on incrementally adapting TinyML models to meet on-device memory constraints.

In parallel, neural architecture search (NAS) methods have started to incorporate robustness into their objectives. \textit{FlatNAS}~\cite{flatnas} introduces a flatness-based metric inspired by rescaling-invariant formulations~\cite{fantastic-generalisation, entropic-flat}, guiding the search toward architectures that are robust to distribution shifts. Related works such as \textit{ADVRUSH}~\cite{advrush}, \textit{GeNAS}~\cite{genas}, $A^2M$~\cite{a2m}, and \textit{NA-DARTS}~\cite{neighborhood-aware} further explore flatness as a proxy for generalization and resilience, using strategies such as eigenvalue regularization, random noise injection, or architectural diversity.

A smaller set of NAS methods directly address the incremental learning problem. \textit{ENAS-S}~\cite{nas-stability} applies a genetic search that jointly optimizes for accuracy and stability, defined as robustness to perturbations. \textit{CLEAS}~\cite{cleas} introduces neuron-level expansion through a reinforcement learning controller that isolates old weights before expanding the model. \textit{Learn-to-Grow}~\cite{learning-to-grow} leverages DARTS to adapt architectures over time and achieves competitive results. \textit{ArchCraft}~\cite{archcraft} is a framework
that integrates evolutionary algorithms with IL. This approach
focuses on optimizing network architectures continuously
while preserving previously acquired knowledge. Lastly, \textit{PCL} ~\cite{pcl} proposes a NAS method to design a population of task-specific models rather than a single robust model for IL.

A recent review~\cite{nas-continual-learning} highlights several pitfalls in current NAS-based IL approaches, namely, redundant retraining, unchecked model growth, and overreliance on replay buffers. SEAL addresses these challenges directly: it avoids retriggering NAS from scratch for each task, expands the architecture only when needed, and removes the dependence on explicit data storage. Furthermore, SEAL exhibits key characteristics desired in IL NAS methods: it supports few-shot learning through cross-task knowledge transfer, maintains a dynamic yet bounded architecture, and introduces mechanisms for graceful forgetting, enabling robustness and scalability across long task sequences. Additionally, another review has identified plasticity-stability balancing as one of the central open challenges for deployable NAS systems, highlighting methods such as SEAL, ArchCraft, and CLEAS as representative approaches \cite{hercules}.


\section{Technical background}
\label{sec:background}

\subsection{Neural Architecture Search}
Neural Architecture Search (NAS) methods can be categorized along three key dimensions~\cite{nas_survey_components}.

The search space defines how architectures are represented. Entire-structured spaces model networks layer by layer, while cell-based spaces, inspired by hand-crafted designs like MobileNets~\cite{mobilenets}, group layers into reusable blocks or cells. Extensions include hierarchical structures and morphism-based transformations.

The search strategy determines how the space is explored. Common approaches include reinforcement learning~\cite{rlsurvey}, gradient-based methods~\cite{liu_darts_2019, zodarts+, zodarts++}, diffusion models \cite{diffusionNAS}, bayesian sampling \cite{squad},  Large Language Models \cite{zeinaty2025}, and evolutionary algorithms~\cite{gambella2023edanas, nachos}. Genetic strategies, especially NSGA-II~\cite{nsganetv2,cnas}, are widely adopted for multi-objective optimization.

The performance estimation strategy aims at evaluating NNs without full training, which is computationally infeasible. A common approach is weight sharing, exemplified by Once-For-All (OFA)\cite{ofa}, which trains a supernet once and evaluates candidates by fine-tuning its weights. Another strategy uses surrogate models like Gaussian Processes or Radial Basis Functions. For example, MSuNAS\cite{nsganetv2} employs adaptive-switching to select the most accurate predictor among four surrogates using Kendall’s Tau correlation in each NAS iteration. Zero-shot NAS methods~\cite{zeroshot} further reduces cost by relying on training-free proxies correlated with final accuracy.

\subsection{Incremental Learning}
\label{sec:incremental-learning}

Incremental Learning (IL) is a machine learning paradigm where models are trained on a sequence of tasks rather than a fixed dataset. Each task comprises a set of observations—possibly from different distributions—used to learn specific actions, depending on the IL setting. IL poses a challenge for deep learning models, which tend to forget previous tasks when learning new ones—a phenomenon known as catastrophic forgetting \cite{types-incremental}. Additionally, their capacity to learn new tasks declines as more training sessions are performed \cite{plasticiy-loss}. \\

Consider a task-agnostic framework, with $\mathcal{D}_{t}$ being a dataset associated to a task $T_{t}$ from a time step $t$. The main objective of an incremental model is to learn the optimal weights $w^*$ from a stream of tasks $\{ T_0, ..., T_t\}$, such that the joint loss function over their datasets is minimized at any given time $t$.

\begin{align}
 w_{t}^* = \min_{w} \sum_{i=0}^{t} \mathcal{L}(D^{*}_{i}|w) 
 \end{align}

In this work, we address a problem of data-incremental learning where disjointed dataset samples with same labels arrive
sequentially and are not retained in memory for the learner
to revisit in the future. Just to mention,  other relevant types of settings are domain-incremental, class-incremental, and task-incremental learning \cite{types-incremental, continual-survey}.

IL performance is typically evaluated across three key aspects: overall performance on all tasks, stability (retention of past knowledge), and plasticity (adaptability to new tasks)~\cite{continual-survey}.

Accuracy remains the standard metric. In IL, we report both average accuracy and average incremental accuracy~\cite{metrics_acc, continual-survey}. The former reflects task performance at a given time, while the latter summarizes progress across all tasks. Let $a_{ki}$ be the accuracy on task $i$ after learning task $k$ ($k \ge i$):

\begin{align}
\text{Average Accuracy} \quad AA_k &= \frac{1}{k} \sum_{i=0}^{k} a_{ki} \\
\text{Average Incremental Accuracy} \quad AIA_k &= \frac{1}{k} \sum_{i=0}^{k} AA_k
\end{align}

Forgetting quantifies knowledge loss, while backward transfer (BWT) measures the impact of new learning on previous tasks~\cite{metrics_acc, bwt-fwt}:

\begin{align}
\text{Forgetting} \quad f^{k}_{i} &= \left( \max_{j \in \{ 0, \dots, k-1\}} a_{ji} \right) - a_{ki} \\
\text{Average Forgetting} \quad F_{k} &= \frac{1}{k} \sum_{i=0}^{k} f^{k}_{i} \\
\text{Backward Transfer} \quad BWT_{k} &= \frac{1}{k} \sum_{i=0}^{k} \left( a_{ki} - a_{ii} \right)
\end{align}
Then, $f^{k}_{i}$ is the forgetting of the task $i$ after training on $k$ tasks.

In data-incremental settings—where tasks are highly related—the max accuracy for a task may not occur immediately after learning it, making average forgetting a more robust metric.

Forward transfer (FWT) approximates plasticity by measuring the influence of past learning on future tasks~\cite{bwt-fwt}:

\begin{align}
FWT_{k} = \dfrac{1}{k}\sum^{k}_{i=1} (a_{(i-1)i} - \bar{b}_{i})
\end{align}

where $\bar{b}_{i}$ is the accuracy of a randomly initialized model trained only on task $i$. FWT thus reflects the performance gain due to prior experience.

\section{The proposed SEAL}
\label{sec:seal}

This section, introducing and detailing SEAL, is organized as follows:
Section~\ref{subsec:problem_formulation} provides the problem formulation,
Section~\ref{subsec:overallview} provides an overall description of SEAL,
Section~\ref{subsec:incremental_models} describes the incremental setting addressed by our framework,
Section~\ref{subsec:expansion_operator} explains the expansion operator used to expand the candidate NNs,
Section~\ref{subsec:cross-distillation} provides the details about the training of the candidate NNs,
Section~\ref{subsec:metrics} presents the figure of merits accounting for accuracy, the number of parameters and flatness of the incremental model.

\subsection{Problem formulation}
\label{subsec:problem_formulation}

SEAL addresses the problem of selecting a NN architecture and its expansion in a data-incremental learning setting by simultaneously optimizing incremental accuracy and robustness to model expansion.
SEAL can be defined as the following optimization problem:
\begin{align}
\label{eq:incremental_problem}
    \text{minimize }  &\mathcal{G} \left ( 
    \mathcal{F}(\bar{x},\vec{d}, w), \mathcal{H}(\bar{x},\vec{d}, \sigma) \right ) 
    \nonumber \\
    \text{s. t. } & \sum_{i=0} \vec{d}_i > 0, \\
    & \bar{x} \in \Omega_{\bar{x}} \nonumber
\end{align}

where $\mathcal{G}$ is a multi-objective optimization function, $\vec{d}$ is the expansion operator, $\bar{x}$ and $\Omega_{\bar{x}}$ represent a candidate NN architecture and the search space of the NN exploration, respectively; the novel metrics $\mathcal{F}(\bar{x},\vec{d}, w)$ and $\mathcal{H}(\bar{x},\vec{d}, \sigma)$ calculate the accuracy in the incremental setting and the robustness to model expansion respectively, and will be introduced in Section \ref{subsec:metrics}. The constraint on the expansion vector ensures that at least one expansion is applied. The optimization problem outlined in Eq. \ref{eq:incremental_problem} is addressed by the SEAL framework, which is detailed in what follows.

\begin{figure*}[ht!]
    \centering
    \includegraphics[width=0.7\linewidth]{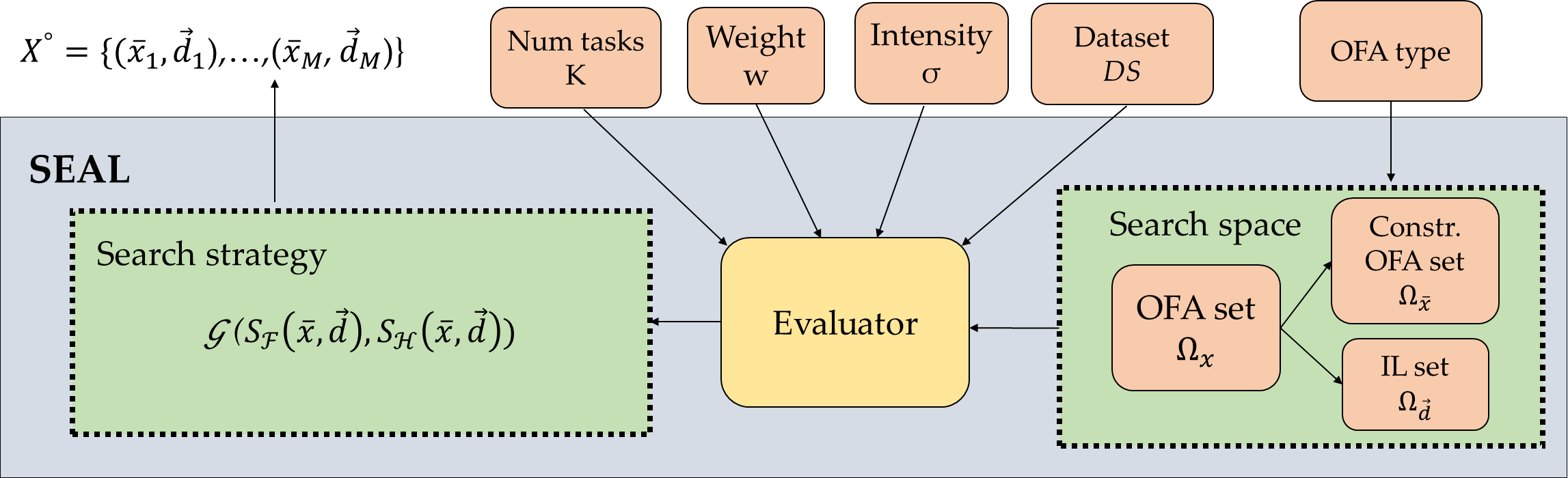}
    \caption{SEAL diagram of the main components. The output of the NAS is a candidate
set of tuples, each containing a NN architecture and its corresponding expansion vector.}
    \label{fig:seal-diagram}
\end{figure*}

\subsection{The overall view of SEAL}
\label{subsec:overallview}

The SEAL framework, illustrated in Fig. \ref{fig:seal-diagram}, operates as follows. It receives a dataset $\mathcal{DS}$, which includes a training set for training candidate networks and a validation set for their validation. The framework utilizes a type of OFA supernet to select a set of candidate networks $\Omega_{x}$, defining the \textit{Search Space} of the NAS. It receives a $K$ parameter defining the number of splits of the dataset for the data-incremental setting. The inputs $w$ and $\sigma$ are parameters used by the first and second objectives of the NAS, respectively.

The \textit{Search Space} module identifies a constrained search space $\Omega_{\bar{x}}$ inside the whole search space to sample only the candidate networks used as base models for the expansion. Additionally, it determines $\Omega_{\vec{d}}$, which is the search space for the possible expansions, ensuring at least one expansion. The pretrained weights of the expanded networks are sampled from the original search space $\Omega_{x}$.

The Evaluator module receives a NN architecture from the set $\Omega_{\bar{x}}$ and trains it by using the $\mathcal{DS}$ dataset following the data-incremental setting with K splits. Subsequently, it assesses incremental accuracy and robustness. The resulting output is an \textit{archive}, i.e., a collection of tuples $\langle \bar{x}, \vec{d}, \mathcal{F}(\bar{x}, \vec{d}, w), \mathcal{H}(\bar{x},\vec{d}, \sigma) \rangle$, representing the NN architecture, the expansion vector, and their corresponding novel figure of merits, i.e., $\mathcal{F}(\bar{x}, \vec{d}, w)$ for the incremental accuracy, and $\mathcal{H}(\bar{x},\vec{d}, \sigma)$ for the robustness to model expansion. Initially, a representative subset of the $\Omega_{\bar{x}}$ is sampled before starting the search. The number $N_{start}$ of selected NN architectures is determined by the user and it is typically set to 100. Therefore, the initial size of the archive is equal to $N_{start}$.

In SEAL, the optimization problem outlined in Eq. \ref{eq:incremental_problem} is reformulated as follows:

\begin{align}
\label{eq:seal_problem}
    \text{minimize }  &\mathcal{G} \left ( 
    \mathcal{S_F}(\bar{x},\vec{d}), \mathcal{S_H}(\bar{x},\vec{d}) \right ) 
    \nonumber \\
    \text{s. t. } & \sum_{i=0} \vec{d}_i > 0, \\
    & \bar{x} \in \Omega_{\bar{x}} \nonumber
\end{align}

where $\mathcal{S_F}(\bar{x},\vec{d})$ is the predicted value of $\mathcal{F}(\bar{x},\vec{d})$ as estimated by using the incremental accuracy surrogate model, and $\mathcal{S_H}(\bar{x},\vec{d})$ is the predicted value of $\mathcal{H}(\bar{x},\vec{d})$ by using the robustness surrogate model.

The core of SEAL is the \textit{Search Strategy} module. It adopts the NSGA-II genetic algorithm to solve the bi-objective problem introduced in Eq. \ref{eq:seal_problem}, by optimizing the objectives $\mathcal{S_F}(\bar{x},\vec{d})$ and $\mathcal{S_H}(\bar{x},\vec{d})$. The search process is iterative: at each iteration, the two surrogate models are chosen by employing a mechanism called \textit{adaptive-switching}, the same method used by MSuNAS, which selects the best surrogate model according to a correlation metric (i.e., Kendall's Tau). Then, a ranking of the candidate NNs, based on $\mathcal{S_F}(\bar{x},\vec{d})$ and $\mathcal{S_H}(\bar{x},\vec{d})$, is computed, and a new set of candidates is obtained by the genetic algorithm and forwarded to the \textit{Evaluator}. The \textit{Evaluator} updates the \textit{archive}, which becomes available for evaluation in the next iteration.
At the end of the search, SEAL returns the set of the $M$ NN architectures and expansions $X^\circ = \{ (\bar{x}_1, \vec{d}_1) \ldots, (\bar{x}_M,\vec{d}_M) \}$ characterized by the best trade-off among the objectives, where $M$ is a user-specified value.

\subsection{Incremental Setting}
\label{subsec:incremental_models}

To decide when to expand the model, we assess its capacity to learn new data by comparing the current cross-entropy loss to that of previous training iterations. Unlike fixed thresholds used in prior works~\cite{den-continual}, this relative comparison provides a more robust and adaptive criterion, as it reflects how well the model handles newly arrived data based on past performance. Formally, given the weights $W_{t-1}$ and loss $L_{t-1}$ from the previous sample set, the model is expanded if
\begin{align}
1 - \dfrac{ \mathcal{L}(W_{t-1}, \mathcal{D}_{t})}{ \mathcal{L}_{t-1}} < \tau
\end{align}
with the hyperparameter $\tau$ defined as the capacity threshold.

\subsection{Expansion of a candidate NN}
\label{subsec:expansion_operator}

\begin{figure}[h!]
    \centering
    \includegraphics[width=0.7\linewidth]{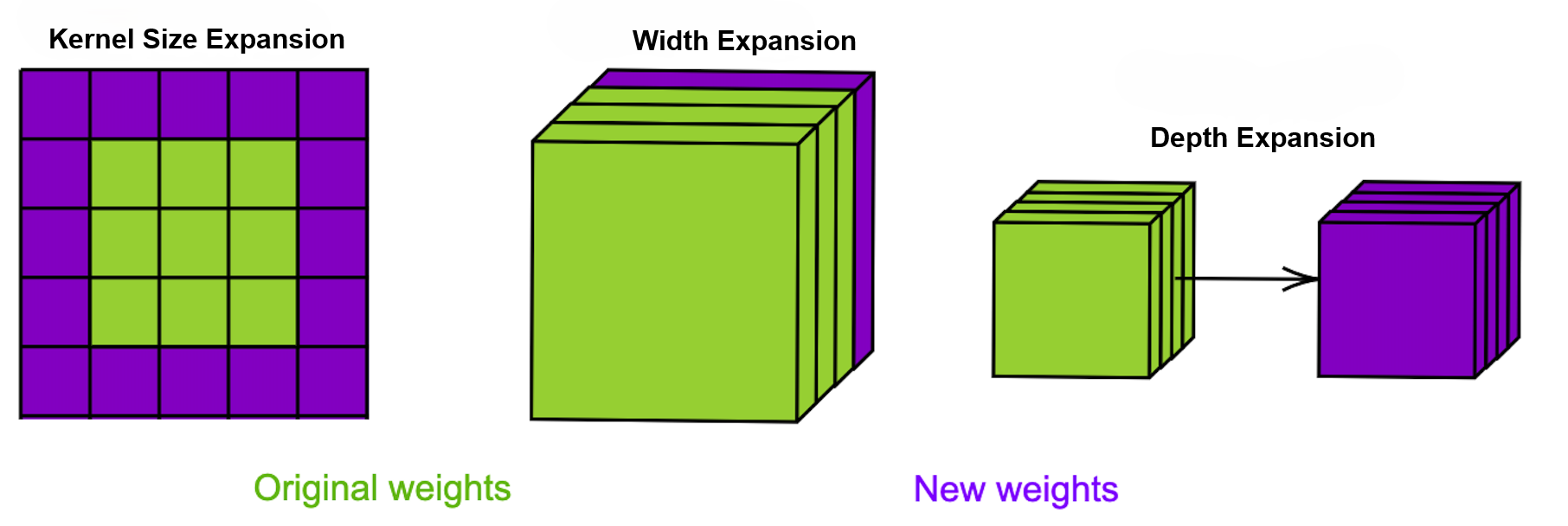}
    \caption{The model expansions performed by SEAL.}
    \label{fig:model-expansions}
\end{figure}

In an OFA setup, we reuse the weights and structure of the pre-trained OFA supernet as much as possible. To ensure compatibility, the constrained OFA set $\Omega_{\bar{x}}$ restricts expansions to produce valid sub-networks within the OFA space. Each expansion inherits weights from the original model and applies changes to the first available OFA-compliant block, starting from the last layers. The expansion is defined by a 3-bit vector $\vec{d}$ indicating changes in depth, width, and kernel size. Figure~\ref{fig:model-expansions} illustrates these operations.

\subsection{Training with cross-distillation}
\label{subsec:cross-distillation}

Expanding the model and so increasing model plasticity to learn new samples, nevertheless, may introduce interference in previously learned representations. To avoid this issue, after each expansion, the model is trained with a \textit{cross-distillation} loss. This forces the added weights to keep a functional representation close to the one before the expansion, reducing catastrophic forgetting \cite{lwf, dynamax}. The distillation loss uses the \textit{Kullback-Leiber Divergence Loss (KL)} on soft targets created by passing the new samples through the old model. The final loss for the expansion follows:
\begin{align}
 \mathcal{L}_{CD} = (1 - \alpha) \mathcal{L}_{CE}(W_{t}, \mathcal{D}_{t}) + \alpha \mathcal{L}_{KL}(W_{t-1}, D_{t})
\end{align}


\subsection{The novel figure of merits of SEAL}
\label{subsec:metrics}
The framework casts a multi-objective optimization problem, with the first objective being a relation between the \textit{average accuracy} at the last incremental sample set and the \textit{number of parameters} of the model. We define the first objective as follows: 
\begin{align}
\mathcal{F}(\bar{x}, \vec{d}, w) = (1 - \mathrm{AA_{K}}(\bar{x}, \vec{d})) \cdot \texttt{size}(\bar{x},\vec{d})^w, \ \ w = 0.07
\end{align}
where $\texttt{size}(\bar{x},\vec{d})$ is the number of parameters of the expanded model.

For the second objective, we use a rescaling-invariant flatness metric, a similar measurement used in \cite{flatnas}. In this definition, the flatness is measured as the change in performance when perturbing the weights under multiplicative noise. Given $\sigma$ the noise intensity, the final flatness is evaluated using a neighborhood of $N$ (i.e., sampling N perturbations):
\begin{align}
\Delta \mathrm{ACC}_{K} &= \left\| \mathrm{ACC}_{\mathcal{D}_{K}}(W_K) - \mathrm{ACC}_{\mathcal{D}_{K}}(W_K + \sigma \cdot z_{i} \odot W_K) \right\| \\
\mathcal{H}(\bar{x}, \vec{d}, \sigma) &= \frac{1}{N} \sum_{i=1}^{N} \frac{ \mathrm{ACC}_{\mathcal{D}_{K}}(W_K) - \Delta \mathrm{ACC}_{K}(W_K, \sigma)}{\mathrm{ACC}_{\mathcal{D}_{K}}(W_K)}
\end{align}
where $\odot$ denotes the element-wise product, $z_i$ is a perturbation sampled by a Gaussian distribution, $W_K$ are the weights of the NN at the last task. 

\section{Experimental Results}
\label{sec:experimental_results}

\subsection{Datasets}

We evaluate our methods on three popular benchmark datasets for NAS~\cite{nasbench201}: CIFAR-10~\cite{cifar}, CIFAR-100~\cite{cifar}, and ImageNet16-120\cite{imagenet16}. CIFAR-10 and CIFAR-100 consist of $32 \times 32$ color images, representing 10 and 100 classes, respectively. Each dataset contains 60000 images split into 50000 training images and 10000 testing images. ImageNet16-120 is a downsized version of the ImageNet dataset \cite{imagenet-dataset} tailored for NAS evaluation, comprising $16 \times 16$ color images across 120 object classes.

\subsection{Analysis of the results}

For all experiments, we use a supernet based on \textit{MobileNetV3} \cite{mobilenets} provided by the OFA library, where the neural architecture search (NAS) space $\Omega_{\bar{x}}$ includes the following hyperparameters: input resolution, network depth, width, and kernel size. The evaluation of candidate architectures, the configuration of the genetic algorithm, and the decision to run SEAL for 30 epochs while generating 8 new candidates per iteration follow the experimental setting of FlatNAS \cite{flatnas}. SEAL has been executed with 5 different seeds on all the datasets, and we report the mean and the standard deviation of these runs.
The threshold value $\tau$ used for triggering architecture expansion is set to 0.2 for CIFAR-10, and 0.13 for the other datasets.
Although the most relevant metric is the $AA$ computed after the last sample set training, all the metrics are tracked across the training. 
Tables \ref{tab:results-objectives-5} and \ref{tab:mono-vs-multi} report the average accuracy, average forgetting, and forward transfer
values. The average incremental accuracy and the backward transfer are not presented as their values
directly relate to the average accuracy and the average forgetting, respectively. Finally, to evaluate the models post-search, we aim at simulating a real-case scenario. As in incremental learning
we are interested in the change in performance for the training samples. Instead of evaluating the model in a
completely new dataset we augment the original training set with more samples, coming from data splits not
available during the NAS. In this way, we are able to accurately measure how the model and expansion policy
will perform in the future. These experiments were performed under a regime of K=5 sequential splits of
disjoint samples, where the model was trained with 3 epochs per task. We emphasize that the model is expanded only if triggered by the condition presented in \ref{subsec:incremental_models}, checked on every new task instead of growing at any task to retain the model complexity. We conducted experiments on a workstation with an NVIDIA A40 GPU (46 GB VRAM), Intel Xeon CPU, and 128 GB RAM, running Ubuntu 20.04 LTS. The implementation uses PyTorch 2.0.1, and the SEAL search cost is around 2 GPU days. 

\begin{table}[h!]

    \centering
    \caption{Comparison between our searched NN architecture and expansion policy
against standard usage of baseline methods}
\scalebox{0.85}{
    \label{tab:results-objectives-5}\resizebox{\columnwidth}{!}{%
    \begin{tabular}{|l|c|c|c|}
        \hline
        \multicolumn{4}{|c|}{\textbf{CIFAR-10@5}} \\
        \hline
        \textbf{Method} & ACC & Forgetting & FWT \\
        \hline
        Naive & 93.18 $(\pm 0.05)$ & 1.41 $(\pm 0.50)$ & 0.68 $(\pm 0.27)$\\
        Joint & 94.85 $(\pm 0.15)$ & - & - \\
        \hline
        EWC & 93.87 $(\pm 0.61)$ & 1.13 $(\pm 0.30)$ & 1.33 $(\pm 0.45)$ \\
        SI &  94.31 $(\pm 0.37)$ & 0.71 $(\pm 0.43)$ & 1.44 $(\pm 0.02)$ \\
        LWF & 92.49 $(\pm 2.16)$ & 2.18 $(\pm 2.27)$ & 1.22 $(\pm 0.84)$ \\
        \hline
        Ours & 95.35 $(\pm 0.18)$ & 1.76 $(\pm 0.41)$ & 0.18 $(\pm 0.19)$\\
        \hline
        \multicolumn{4}{c}{} \\
        
        \hline 
        \multicolumn{4}{|c|}{\textbf{CIFAR-100@5}} \\
        \hline
        \textbf{Method} & ACC & Forgetting & FWT \\
        \hline
        Naive & 81.32 $(\pm 0.85)$ & 7.65 $(\pm 0.66)$ & -0.45 $(\pm 1.21)$ \\
        Joint & 84.71 $(\pm 0.65)$ & - & - \\
        \hline
        EWC & 82.38 $(\pm 0.37)$ & 6.94 $(\pm 0.40)$ & 0.58 $(\pm 0.60)$ \\
        SI &  82.55 $(\pm 0.39)$ & 5.89 $(\pm 0.48)$ & -0.53 $(\pm 0.75)$\\
        LWF & 83.21 $(\pm 0.25)$ & 5.50 $(\pm 0.14)$ & 2.65 $(\pm 0.27)$ \\
        \hline
        Ours &83.94 $(\pm 0.33)$ & 4.81 $(\pm 0.37)$ & 1.59 $(\pm 0.25)$\\
        \hline
        \multicolumn{4}{c}{} \\

        \hline
        \multicolumn{4}{|c|}{\textbf{ImageNet-16@5}} \\
        \hline
        \textbf{Method} & ACC & Forgetting & FWT \\
        \hline
        Naive & 51.71 $(\pm 0.15)$ & 5.54 $(\pm 0.35)$ & 1.00 $(\pm 0.37)$\\
        Joint & 51.84 $(\pm 1.30)$ & - & - \\
        \hline
        EWC &51.75 $(\pm 0.03)$ & 5.50 $(\pm 0.65)$ & 2.00 $(\pm 1.14)$ \\
        SI & 51.39 $(\pm 0.12)$ & 5.73 $(\pm 0.44)$ & 0.83 $(\pm 0.41)$ \\
        LWF & 53.02 $(\pm 0.10)$ & 4.10 $(\pm 0.15)$ & 3.27 $(\pm 0.24)$\\
        \hline
        Ours & 52.45 $(\pm 0.13)$ & 3.23 $(\pm 0.24)$ & 6.98 $(\pm 0.18)$ \\
        \hline
        
    \end{tabular}
    }
    }
    
\end{table}

\begin{table*}[h!]
    \centering
        \caption{Multi-objective vs single-objective NAS search results for a 5-splits data-incremental problem. Note: * denotes that the selected models coincide.}
    \label{tab:mono-vs-multi}
    \begin{tabular}{|l|c|c|c|c|c|}
        \hline
        \multicolumn{6}{|c|}{\textbf{CIFAR-10@5}} \\
        \hline
        \textbf{Selection} & Params & Flatness & ACC & Forgetting & FWT \\
        \hline
        SingleObj & 3.98M $(\pm 0.63)$ & 73.61 $(\pm 1.82)$ & 94.82 $(\pm 0.57)$ & 2.09 $(\pm 1.00)$ & 0.55 $(\pm 0.74)$ \\
        MultiObj-Acc & 3.92M $(\pm 0.50)$ & 75.05 $(\pm 1.14)$ & 95.35 $(\pm 0.18)$ & 1.76 $(\pm 0.41)$ & 0.18 $(\pm 0.19)$\\
        MultiObj-Flat & 3.79M $(\pm 0.40)$ & 82.09 $(\pm 2.13)$ & 94.53 $(\pm 0.51)$ & 1.37 $(\pm 0.54)$ & 0.75 $(\pm 0.34)$ \\
        MultiObj-Trade & 3.65M $(\pm 0.15)$ & 78.19 $(\pm 3.33)$ & 94.89 $(\pm 0.68)$ & 1.89 $(\pm 0.56)$ & 0.25 $(\pm 0.25)$ \\
        \hline
        \multicolumn{6}{c}{} \\
        
        \hline 
        \multicolumn{6}{|c|}{\textbf{CIFAR-100@5}} \\
        \hline
        \textbf{Selection} & Params & Flatness & ACC & Forgetting & FWT \\
        \hline
        SingleObj & 4.37M $(\pm 0.06)$ & 52.57 $(\pm 1.84)$ & 83.01 $(\pm 1.20)$ & 5.72 $(\pm 1.28)$ & 0.83 $(\pm 1.02)$ \\
        MultiObj-Acc* & 4.09M $(\pm 0.21)$ & 57.88 $(\pm 3.48)$ & 83.94 $(\pm 0.33)$ & 4.81 $(\pm 0.37)$ & 1.59 $(\pm 0.25)$  \\
        MultiObj-Flat & 3.85M $(\pm 0.02)$ & 58.49 $(\pm 2.87)$ & 83.61 $(\pm 0.01)$ & 5.14 $(\pm 0.04)$ & 1.12 $(\pm 0.22)$ \\
        MultiObj-Trade* & 4.09M $(\pm 0.21)$ & 57.88 $(\pm 3.48)$ & 83.94 $(\pm 0.33)$ & 4.81 $(\pm 0.37)$ & 1.59 $(\pm 0.25)$ \\
        \hline
        \multicolumn{6}{c}{} \\

        \hline
        \multicolumn{6}{|c|}{\textbf{ImageNet-16@5}} \\
        \hline
        \textbf{Selection} & Params & Flatness & ACC & Forgetting & FWT \\
        \hline
        SingleObj* & 4.24M $(\pm 0.51)$ & 45.42 $(\pm 1.82)$ & 52.31 $(\pm 0.01)$ & 3.35 $(\pm 0.28)$ & 6.91 $(\pm 0.17)$ \\
        MultiObj-Acc* & 4.24M $(\pm 0.51)$ & 45.42 $(\pm 1.82)$ & 52.31 $(\pm 0.01)$ & 3.35 $(\pm 0.28)$ & 6.91 $(\pm 0.17)$ \\
        MultiObj-Flat & 3.55M $(\pm 0.06)$ & 56.16$(\pm 1.89)$ & 50.40 $(\pm 0.20)$ & 2.29$(\pm 0.22)$ & 4.21 $(\pm 0.34)$\\
        MultiObj-Trade & 4.06M $(\pm 0.36)$ & 49.98 $(\pm 2.59)$ & 51.51 $(\pm 0.29)$ & 2.24 $(\pm 0.26)$& 4.33$(\pm 0.25)$ \\
        \hline
    \end{tabular}

\end{table*}

\begin{figure*}[h!]
    \centering
    \includegraphics[width=0.9\linewidth]{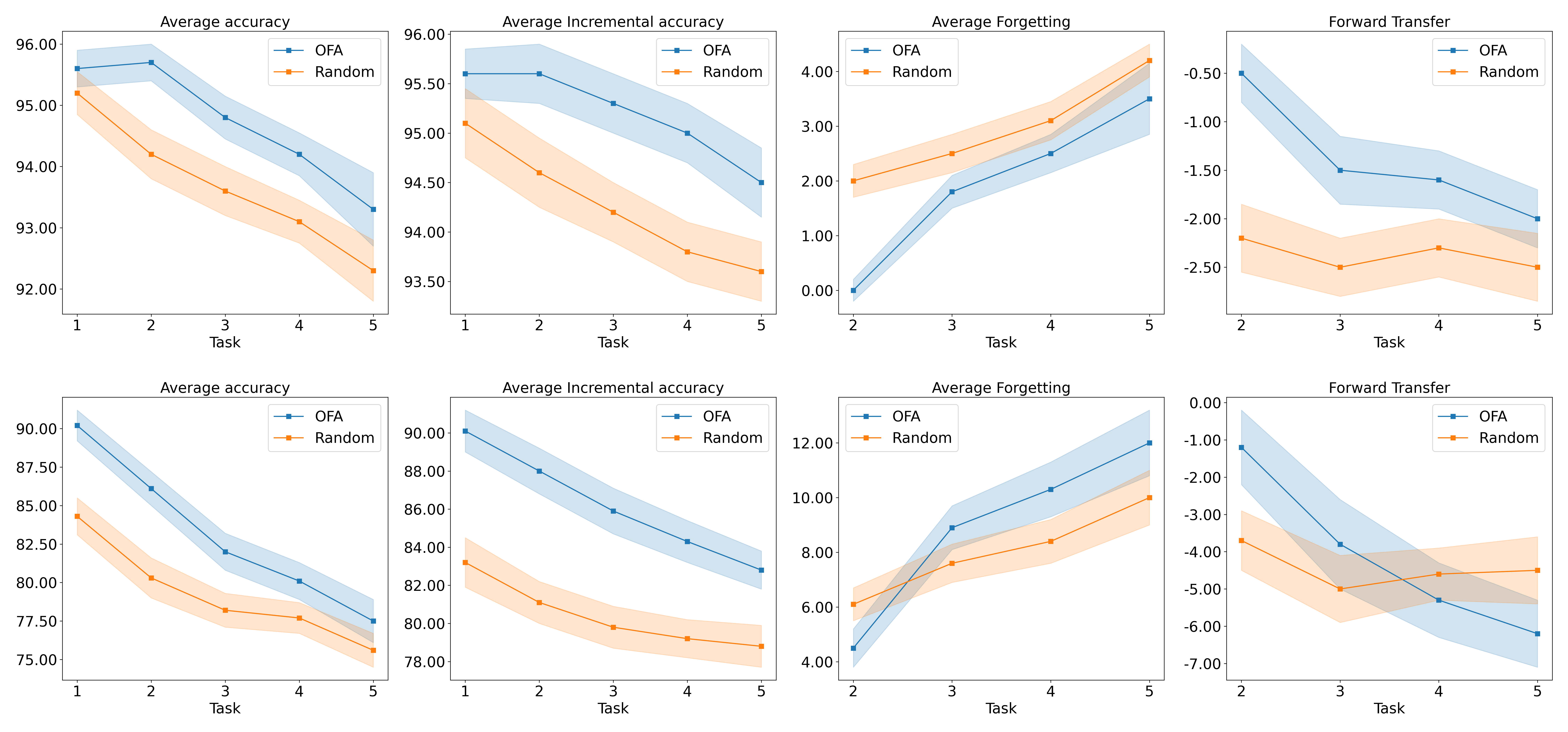}
    \caption{Performance distribution for our IL problem. Expansion starts from random weights vs OFA-initialized weights on the CIFAR-10 and CIFAR-100 datasets.}
    \label{fig:random-vs-ofa}
\end{figure*}

\begin{figure*}[h!]
    \centering
    \includegraphics[width=0.9\linewidth]{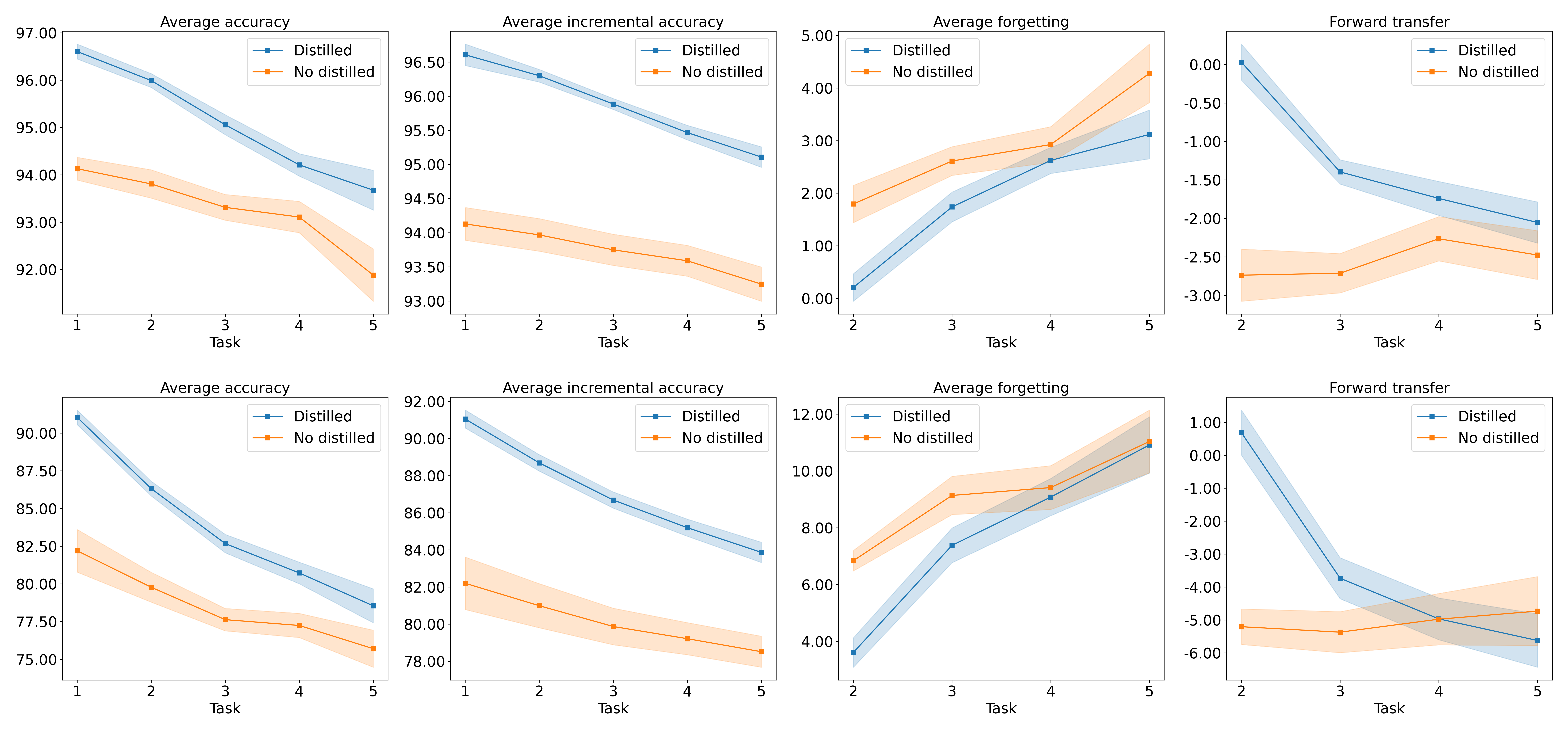}
    \caption{Performance distribution when models are trained with the distillation loss on CIFAR-10 and CIFAR-100 datasets.}
    \label{fig:distillation-loss}
\end{figure*}

Regarding the baselines, we followed the benchmarking done in \cite{types-incremental} and selected models that suited our constraints and our data-incremental setting. Based on these criteria, we selected EWC \cite{ewc}, LWF \cite{lwf}, and SI \cite{si-baseline}— being straightforward yet state-of-the-art methods across a range of continual learning settings. 
To strengthen the baselines, we included the \textit{naive} and the \textit{joint-training} benchmark cases, common baselines in IL. The \textit{naive} method is a model trained in an incremental way without any additional constraint, it shows the performance degradation of a model just by being trained in a continual setting. On the other hand, the \textit{joint-training} refers to a model trained on the whole dataset at once, as if it were a static dataset. All the benchmarks have been trained with 5 different seeds, and we report the mean and the standard deviation of these runs. 

In Table \ref{tab:results-objectives-5}, we show our main result, i.e., that our SEAL models are competitive with the other baselines in every metric on each dataset. Notably, the \textit{naive} approach sometimes achieves high accuracy—especially on CIFAR-10—when tasks are similar and do not require strong regularization. However, while SEAL outperforms \textit{naive} in overall performance, it also shows increased forgetting and lower forward transfer on CIFAR-10, potentially indicating overfitting on the latest tasks. Finally, \textit{joint-training} does not always yield the highest accuracy, likely due to task interference in heterogeneous datasets. 

To validate the statistical significance of our results, we performed the Friedman statistical test by using each metric of the multiple runs over five seeds of all our tested methods, for each dataset and reported results in Table \ref{tab:friedman_results}. The results shows that there are statistic differences among the methods, hence the competitive performance of SEAL is statistically validated.

\begin{table}[h!]
\centering
\caption{Friedman test results for each dataset and objective. 
The higher $\chi^2$, the more different the results of the methods based on their rankings. 
A smaller p-value (typically $< 0.05$) indicates more statistically significant differences.}
\label{tab:friedman_results}
\begin{tabular}{lcc}
\toprule
\textbf{Dataset \& Objective} & \textbf{Chi-squared ($\chi^2$)} & \textbf{p-value} \\
\midrule
CIFAR-10 (ACC)        & 24.54  & 0.00017 \\
CIFAR-10 (Forgetting) & 21.34  & 0.00070 \\
CIFAR-10 (FWT)        & 20.77  & 0.00089 \\
CIFAR-100 (ACC)       & 22.94  & 0.00035 \\
CIFAR-100 (Forgetting)& 22.60  & 0.00040 \\
CIFAR-100 (FWT)       & 20.77  & 0.00089 \\
ImageNet-16 (ACC)     & 22.26  & 0.00047 \\
ImageNet-16 (Forgetting) & 24.31 & 0.00019 \\
ImageNet-16 (FWT)     & 23.17  & 0.00031 \\
\bottomrule
\end{tabular}
\end{table}

We compare performance between single-objective optimization using the mono-objective genetic algorithm Differential Evolution \cite{differentiableevolution}, and multi-objective optimization incorporating the flatness objective. Table \ref{tab:mono-vs-multi} shows metrics for four models, each row aggregating the searched model from five NAS runs: the searched single-objective model (accuracy) and the searched multi-objective models by accuracy, trade-off, and flatness. The results validate the superiority of the multi-objective problem and the use of flatness for incremental learning. An additional insight is how increasing the flatness seems to correlate with smaller models. This may suggest that flatten-minima models may require fewer parameters to fully learn the problem, an important insight for constrained-resources environments were NAS are commonly used.

Additionally, we show in Fig. \ref{fig:random-vs-ofa} the performance differences between initializing the new sections of the model from random
weights, without using the OFA set, and initializing the weights from the OFA. It is clear that initializing from the OFA set has a beneficial effect over the
model accuracy and forgetting, although it is relevant to notice that this effects seems to vanish with increases
in data complexity.

Furthermore, we validate the usage of the cross-distillation loss for training the expanded models. From Fig. \ref{fig:distillation-loss}, there is a clear performance increase by using the distillation approach. Not only the accuracies achieve higher values, but the forgetting and knowledge transfer metrics also improve. 

\section{Conclusions}
\label{sec:conclusions}

This paper introduced SEAL, the first NAS framework to jointly optimize both architecture and expansion policy for DNNs. Our results demonstrate that SEAL reduces forgetting and improves accuracy, while selectively allocating additional capacity only when required. While designed for data-incremental learning, SEAL can be extended to other IL settings. A promising direction for future work is to strengthen NAS’s role in dynamic model growth. In this study, some decisions were fixed under the assumption that architectures generalize across data splits—an assumption that may not hold in domain- or class-incremental settings. Incorporating a capacity-aware control signal (e.g., a threshold-bit) into the NAS process could enable more adaptive expansion decisions.

\bibliographystyle{IEEEtran}
\bibliography{main}


\clearpage

\end{document}